\documentclass[10pt,twocolumn,letterpaper]{article}

\usepackage[pagenumbers]{wacv} %

\usepackage{graphicx}
\usepackage{amsmath}
\usepackage{amssymb}
\usepackage{booktabs}
\usepackage{algorithm}
\usepackage{algorithmic}
\usepackage{amsthm}
\usepackage{proof}
\usepackage{microtype}
\usepackage{makecell}
\usepackage{multirow}
\usepackage{nicefrac}
\usepackage{tabularx}
\usepackage[table]{xcolor}
\usepackage{url}
\usepackage{xcolor}
\usepackage{lipsum}
\usepackage{enumitem}
\definecolor{Gray}{gray}{0.9}

\newcommand*{\ours}[0]{SpotDiffusion}

\usepackage[pagebackref,breaklinks,colorlinks]{hyperref}

\usepackage[capitalize]{cleveref}
\crefname{section}{Sec.}{Secs.}
\Crefname{section}{Section}{Sections}
\Crefname{table}{Table}{Tables}
\crefname{table}{Tab.}{Tabs.}
\crefname{figure}{Fig.}{Figs.}

\def\wacvPaperID{1459} %
\def\confName{WACV}
\def\confYear{2025}

\begin{document}

\title{SpotDiffusion: A Fast Approach For Seamless Panorama Generation Over Time}

\author{Stanislav Frolov \hspace{2em} Brian B. Moser \hspace{2em} Andreas Dengel\\
\\
German Research Center for Artificial Intelligence (DFKI), Germany \\
RPTU Kaiserslautern-Landau, Germany \\
\texttt{first.last@dfki.de}
}

\maketitle

\begin{abstract}
Generating high-resolution images with generative models has recently been made widely accessible by leveraging diffusion models pre-trained on large-scale datasets.
Various techniques, such as MultiDiffusion and SyncDiffusion, have further pushed image generation beyond training resolutions, i.e., from square images to panorama, by merging multiple overlapping diffusion paths or employing gradient descent to maintain perceptual coherence.
However, these methods suffer from significant computational inefficiencies due to generating and averaging numerous predictions, which is required in practice to produce high-quality and seamless images.
This work addresses this limitation and presents a novel approach that eliminates the need to generate and average numerous overlapping denoising predictions.
Our method shifts non-overlapping denoising windows over time, ensuring that seams in one timestep are corrected in the next.
This results in coherent, high-resolution images with fewer overall steps. 
We demonstrate the effectiveness of our approach through qualitative and quantitative evaluations, comparing it with MultiDiffusion, SyncDiffusion, and StitchDiffusion.
Our method offers several key benefits, including improved computational efficiency and faster inference times while producing comparable or better image quality.
\href{https://github.com/stanifrolov/spotdiffusion}{Link to code}
\end{abstract}

\section{Introduction}
\label{sec:intro}

\begin{figure}[t]
  \centering
  \includegraphics[width=\linewidth]{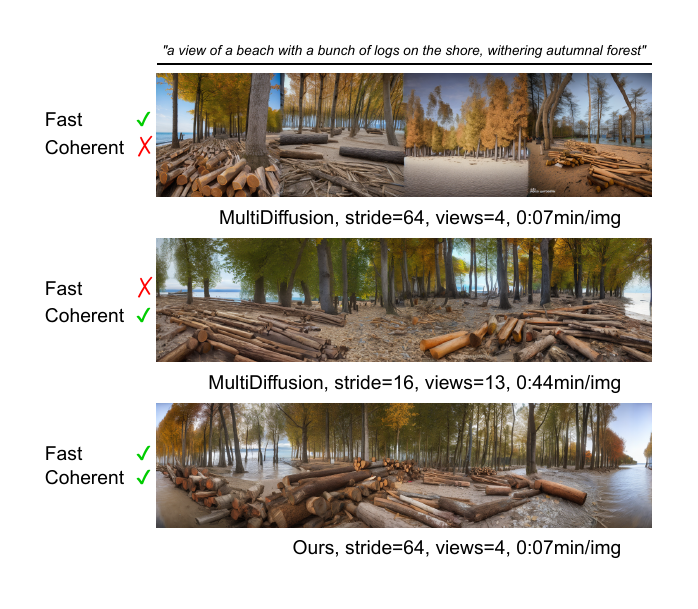}
  \caption{
    MultiDiffusion \cite{bar2023multidiffusion} can produce coherent panorama images by averaging overlapping denoising predictions.
    However, this process introduces computational inefficiencies and requires denoising many patch views.
    Without overlapping views, MultiDiffusion can not produce coherent panoramas.
    We introduce an efficient method for high-resolution panorama generation that eliminates the need for overlapping denoising predictions, resulting in coherent and sharp images without border artifacts.
  }
  \label{fig:teaser} 
\end{figure}

The recent breakthrough of generative diffusion models \cite{dhariwal2021diffusion,ramesh2021zero,rombach2022high} has enabled the generation of high-quality images with high fidelity and diversity from simple text prompts.
These models are typically trained on a large-scale dataset of fixed-size resolution.
Training such models for high-resolution image generation requires substantial computational resources and large datasets.
Therefore, several attempts have been made to repurpose pre-trained diffusion models to exceed the original training image resolution.
When initially trained on lower resolutions, innovative techniques are required to overcome their original constraints (fixed spatial size) and generate higher-resolution images.
Previous methods, such as MultiDiffusion \cite{bar2023multidiffusion} and SyncDiffusion \cite{lee2023syncdiffusion}, have made significant strides in this domain.
MultiDiffusion \cite{bar2023multidiffusion} achieves high resolution by fusing multiple overlapping diffusion paths, which are then averaged to create the final image.
While effective, this technique introduces computational inefficiencies due to the need to generate and average numerous predictions.
SyncDiffusion \cite{lee2023syncdiffusion} builds on this by using gradient descent from a perceptual similarity loss to produce coherent panoramas, yet it still fundamentally relies on the principles of MultiDiffusion.
A major drawback of these current methods is the high computational cost required to create overlapping predictions during denoising and combine them through averaging.
For example, in practice, MultiDiffusion \cite{bar2023multidiffusion} requires an overlap of 75\% between adjacent windows, resulting in a large number of denoising predictions and thus significantly increased computational demands.

In this paper, we present our novel approach, coined SpotDiffusion (\textbf{s}eamless \textbf{p}anorama \textbf{o}ver \textbf{t}ime), that addresses this limitation and offers a more efficient and effective method for high-resolution image generation.
We argue that many of the denoising predictions, like in MultiDiffusion \cite{bar2023multidiffusion}, are redundant and that averaging can decrease the quality of the final image, in addition to incurring significant computational costs.
Furthermore, we identify that the key to producing seamless panoramas is to ensure that the full denoising process is applied uniformly over time across the entire image.
Our key insight is that this can be achieved without overlapping denoising predictions within one timestep by simply shifting the denoising windows over time.
This ensures that any seam in one timestep is corrected in the next, resulting in coherent, high-resolution images with significantly fewer steps.
SpotDiffusion makes several key contributions:
\setlist{nolistsep}
\begin{itemize}[noitemsep]
    \item it introduces a fast method for coherent and sharp panorama image generation without border artifacts.
    \item it eliminates the need for generating numerous overlapping denoising predictions and subsequent averaging, significantly reducing the computational complexity of the denoising process.
    \item it can serve as a drop-in replacement for existing diffusion models that previously relied on the MultiDiffusion mechanism, making it a practical and efficient solution for high-resolution image generation.
\end{itemize}

\section{Related Work}
\label{sec:related_work}
This section provides an overview of the related work in the field of high-resolution image generation, focusing on diffusion models, text-to-image synthesis, and high-resolution montage generation.

\subsection{Diffusion Models}
Diffusion models \cite{sohl2015deep,ho2020denoising, song2019generative,rombach2022high} are powerful generative models.
A key difference between them and earlier generative models, such as GANs \cite{goodfellow2020generative}, is their iterative process.
During training, diffusion models employ a forward and backward process.
The forward process starts from a clean image and iteratively adds small amounts of noise until the image is indistinguishable from random noise.
In the backward process, they learn to generate real-looking data by progressively reverting the forward process, i.e., removing the added noise.
As a result, they can gradually approximate a complex target data distribution from a normal noise distribution \cite{song2020scorebased,dhariwal2021diffusion}.
These models have been shown to produce high-quality images and have been used in various applications, including image generation, inpainting, super-resolution, and across various data modalities, including audio, video, and 3D objects, showcasing their broad applicability \cite{yang2023diffusion,lugmayr2022repaint,moser2024diffusion,evans2024fast,croitoru2023diffusion}.

\subsection{Text-to-Image Synthesis}
Text-to-image \cite{frolov2021adversarial,zhang2023text} synthesis has gained significant attention for its impressive generation performance. 
Diffusion models, particularly latent diffusion models \cite{rombach2022high}, have become popular due to their high-quality generation capabilities.
Stable Diffusion, built upon latent diffusion models \cite{rombach2022high}, has been shown to produce high-resolution images with high fidelity and diversity making it a popular choice for text-to-image synthesis and enabling researchers as well as artists to generate creative images from text prompts. 
Despite their strengths, these models are limited by their training resolution, necessitating innovative techniques to exceed their original training limits and produce high-resolution outputs.

\begin{figure*}[t]
  \centering
  \includegraphics[width=\textwidth]{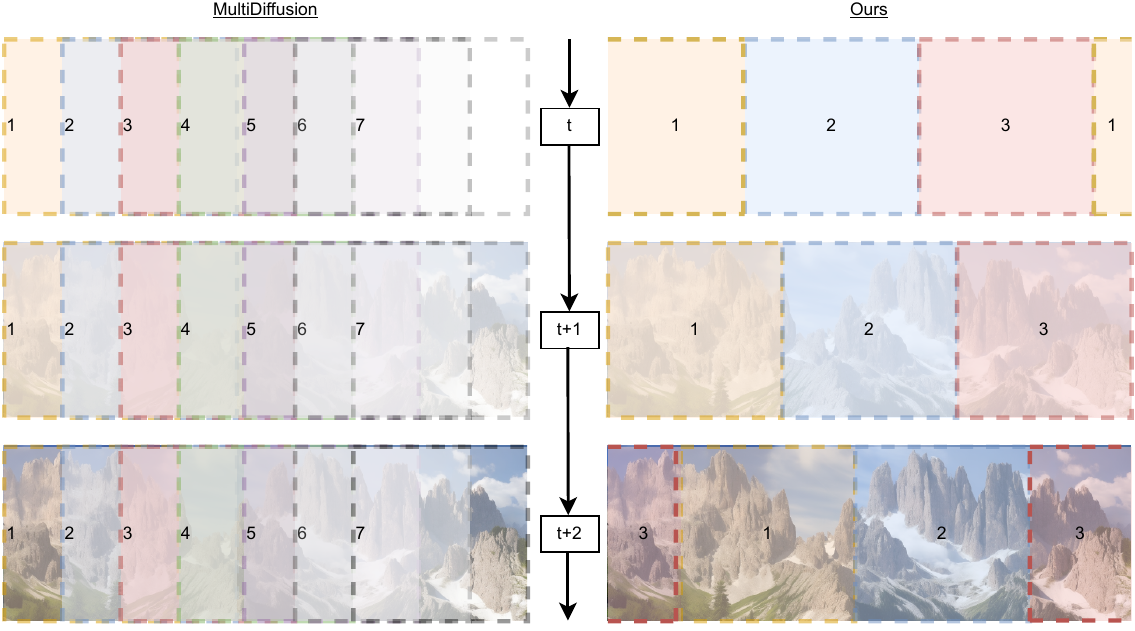}
  \caption{
    MultiDiffusion \cite{bar2023multidiffusion} generates coherent panorama images by averaging overlapping denoising predictions with a stride that is smaller than the denoising window.
    Our method eliminates the need for overlapping denoising predictions and introduces a more efficient shifting denoising method.
    Instead of relying on a fixed denoising path with overlapping views, our method shifts the denoising windows over time, ensuring that seams in one timestep are corrected in the next.
    This results in fast, seamless, high-resolution images with fewer overall steps.
    }
  \label{fig:spotdiff} 
\end{figure*}

\subsection{High-Resolution Image Generation}
High-resolution image synthesis presents significant challenges due to the high-dimensional data and substantial computational resources required.
Previous approaches include training from scratch or fine-tuning, which can be computationally expensive and time-consuming.
Training-free methods like MultiDiffusion \cite{bar2023multidiffusion} repurpose pre-trained models on low-resolution to generate high-resolution images.
To that end, multiple overlapping diffusion paths are fused by averaging the denoising predictions to produce seamless images.
SyncDiffusion \cite{lee2023syncdiffusion} improves upon these methods by synchronizing multiple diffusions through gradient descent, achieving more coherent outputs.
However, these methods introduce a significant computational overhead due to the need for generating and averaging numerous predictions, especially because large overlaps are required to produce seamless images.
Our work aims to address these limitations by introducing a novel approach that eliminates the need for overlapping denoising predictions and subsequent averaging by shifting non-overlapping denoising windows over time.

\section{Methodology}
\label{sec:method}
We propose SpotDiffusion, a new diffusion process for fast panorama image generation.
Contrary to previous methods, SpotDiffusion uses randomly shifted, non-overlapping windows through time to ensure coherent transitions.
This section details the mathematical formulation of our method and presents the algorithm for seamless panorama generation.

\subsection{Preliminaries}
Consider a pre-trained diffusion model \(\Phi\) that operates in image space \(\mathcal{I} \in \mathbb{R}^{W\times H\times C} \) and condition space \(\mathcal{Y}\). Given a noisy image \(I_T \sim \mathcal{N}(\mathbf{0}, \mathbf{I})\) and a condition \(y \in \mathcal{Y}\), the model produces a sequence of images starting from $I_T$ and gradually denoising it towards the clean image \(I_0\):
\begin{equation}
I_T, I_{T-1}, \ldots, I_0 \quad \text{such that} \quad I_{t-1} = \Phi(I_t \mid y)
\end{equation}

Our goal is to generate images $J_T, J_{T-1}, \ldots, J_0$ in a new image space \(\mathcal{J} \in \mathbb{R}^{W'\times H'\times C} \), with $W'\geq W, H'\geq H$, using the same reference model \(\Phi\) without any re- or fine-tuning.
Traditionally, models pre-trained on fixed-size images cannot be directly used to produce images of arbitrary sizes.
To that end, MultiDiffusion \cite{bar2023multidiffusion} addressed this limitation by using a joint diffusion approach where multiple overlapping windows are merged via averaging.
More formally, $n$ mappings $F_i: \mathcal{J}\in \mathbb{R}^{W'\times H'\times C} \to \mathcal{I}\in \mathbb{R}^{W\times H\times C}$ with $i \in \{1,2,...,n\}$ are defined that map (crop) the image space \(\mathcal{J}\) into $n$ images of the original space \(\mathcal{I}\).
The value of $n$, and thus the number of individual image crops, is defined as $n=\frac{W^{\prime}-W}{\omega}+1$, where $\omega$ is the stride between adjacent cropping windows.
With these mappings, the denoising process is applied to each cropped image, and the results are averaged to produce the final image $J_0$.

Notably, the stride $\omega$ is typically chosen such that the cropped windows overlap, ensuring that the denoising process is applied uniformly across the entire image.
In practice, it is common to choose $\omega = \frac{W}{4}$ or even $\omega = \frac{W}{8}$ such that the cropped windows overlap by 75\% or 87.5\%, respectively to produce seamless transitions.
However, this approach dramatically increases the computational complexity of the denoising process, as each overlapping window requires a denoising prediction.
Strides $\omega \ge \frac{W}{2}$ are known to produce visible artifacts and seams in the final image.
In the extreme case of $\omega = W$, the cropped windows are disjoint, and the denoising process is applied independently to each window, resulting in a disjoint image.

\subsection{\ours}
Our goal is to efficiently generate high-resolution, seamless images and speed up the total inference time.
Thus, we aim to eliminate the need for many overlapping denoising windows.
We hypothesize that many of the denoising predictions are redundant and that averaging may decrease the quality or alignment of the final image.
To that end, we identify that the key to producing seamless panoramas is to ensure that the denoising process is applied uniformly across the entire image over time.
We recognize that the static mappings $F_i$ in MultiDiffusion \cite{bar2023multidiffusion} are fixed and do not change over time, which we believe is a critical limitation of the method.
Therefore, in this paper, we propose a shifted windows diffusion approach.
Specifically, we use time-dependent mappings $F_{(i,t)}$ using $s(t)$, which applies random shifts of size $\mathtt{s}(t)$ at every timestep to the mapping such that the entire image is denoised uniformly.
Our core insight is that using non-overlapping windows that shift randomly over time effectively corrects discontinuities at the seams in subsequent time steps.
Given sufficient timesteps, our shifted windows ensure that each pixel is processed through various windows, achieving a uniformly denoised and perfectly coherent image.
See \autoref{fig:spotdiff} for a conceptual illustration of our method and comparison to MultiDiffusion \cite{bar2023multidiffusion}.

More formally, for each timestep \(t\), a shift function with a shift size of \(\mathtt{s}(t)\) is applied to the mappings \(F_{(i, t)}\).
The shift size $\mathtt{s}(t) \sim \mathcal{U}(0, W)$ is drawn from a uniform distribution over the entire mapping sequence, ensuring that each pixel in the image \(\mathcal{J}\) is included in multiple windows over time.
In our method, the stride equals the width of the denoising window $\omega = W$, so the windows are non-overlapping.
To handle cases where a window mapping exceeds the image boundaries, \(F_{(i, t)}\) maps it back to the beginning of the image (wrap-around).
In practice, our method can efficiently be implemented by shifting the input image $J_t$ of the mapping functions $F_i$ with a $\mathtt{translate}$ operator that incorporates a wrap-around, as shown in \autoref{alg:shifted_windows_multidiffusion}.
The denoised image $J_{t-1}$ is obtained by concatenating the denoised windows and reverting the shift by re-applying $\mathtt{translate}$ with $-\mathtt{s}(t)$.
Due to the random shifts, any artifacts or seams produced in one timestep are corrected in subsequent timesteps, leading to a coherent and seamless final image.

The key advantage of our method lies in its ability to produce coherent transitions without the need for overlapping windows, which simplifies the method and significantly reduces the time complexity of the denoising process.
Furthermore, our approach can be easily integrated into existing diffusion models, especially as a drop-in replacement for MultiDiffusion \cite{bar2023multidiffusion}, without the need for retraining or fine-tuning, making it a practical and efficient solution for high-resolution image generation.

\section{Experimental Setup}
\label{sec:experiments}

We choose three models to evaluate our method.
First, we compare our approach with MultiDiffusion \cite{bar2023multidiffusion} using the Stable Diffusion 2.0 backbone.
The Stable Diffusion model operates in a latent space of \(\mathbb{R}^{64 \times 64 \times 4}\) and generates images of \(\mathbb{R}^{512 \times 512 \times 3}\).
We use it to generate panorama images of resolution \(512 \times 2048\) (\(64 \times 256\) in the latent space), where the width is four times the width of the output of Stable Diffusion.
Each window crop in the panorama has an image resolution of \(512 \times 512\).
We compare against MultiDiffusion \cite{bar2023multidiffusion} with various strides along the width in the image space.
In the case of no overlaps, the stride equals the window size of 64 resulting in a constant compute comparison with our method.
Second, we apply our method as a drop-in replacement for the inner MultiDiffusion \cite{bar2023multidiffusion} loop in SyncDiffusion \cite{lee2023syncdiffusion}.
SyncDiffusion \cite{lee2023syncdiffusion} synchronizes multiple diffusions through gradient descent to produce coherent panoramas, but requires subsequent MultiDiffusion \cite{bar2023multidiffusion} to merge all predictions.
We replace it with SpotDiffusion to speed up the image generation process.
Finally, we compare our method with StitchDiffusion \cite{wang2024customizing}, a method for spherical image synthesis, which generates 360-degree panoramas by averaging overlapping denoising predictions.
StitchDiffusion \cite{wang2024customizing} customizes a pre-trained T2I diffusion model for 360-degree panorama synthesis by fine-tuning a Low-Rank Adaptation (LoRA) \cite{hu2021lora} matrices.
Here, we also replace the overlapping denoising predictions with our method to demonstrate the efficiency of our approach and show the qualitative results and image generation time.

\begin{algorithm}[t]
  \caption{SpotDiffusion}
  \label{alg:shifted_windows_multidiffusion}
  \begin{algorithmic}[1]
  \REQUIRE $\Phi$ \COMMENT{pre-trained diffusion model} \\
  \hspace{0.8cm} $\{F_i\}_{i=1}^n$ \COMMENT{non-overlapping mappings} \\
  \hspace{0.8cm} $\{y_i\}_{i=1}^n$ \COMMENT{conditions for each window} \\
  \hspace{0.8cm} $J_T \sim \mathcal{N}(\mathbf{0}, \mathbf{I})$ \COMMENT{noisy initialization} \\
  \hspace{0.8cm} $\mathtt{translate}(\cdot, c)$ \COMMENT{translate by $c$ function}
  \FOR{$t = T, T-1, \ldots, 0$}
      \STATE $\mathtt{s}(t) \sim \mathcal{U}(0, W) $ \COMMENT{sample a random shift size}
      \STATE $\hat{J}_t \leftarrow \mathtt{translate} (J_t, \mathtt{s} (t))$ \COMMENT{shift panorama}
          
      \FOR{each window $i = 1, \ldots, n$}
          \STATE $I_{t-1}^{i} \leftarrow \Phi (F_i(\hat{J}_t) \mid y_i)$ \COMMENT{denoise window}
      \ENDFOR
      \STATE $\hat{J}_{t-1} \leftarrow  \mathtt{concat}(\{I_{t-1}^i\}_{i=1}^n)$ \COMMENT{combine windows}
      \STATE $J_{t-1} \leftarrow \mathtt{translate} (\hat{J}_{t-1}, -\mathtt{s}(t))$ \COMMENT{revert shift}
  \ENDFOR
  \STATE \textbf{return} $J_0$
  \end{algorithmic}
\end{algorithm}
  
\begin{figure*}[t]
\centering
\includegraphics[width=0.97\textwidth]{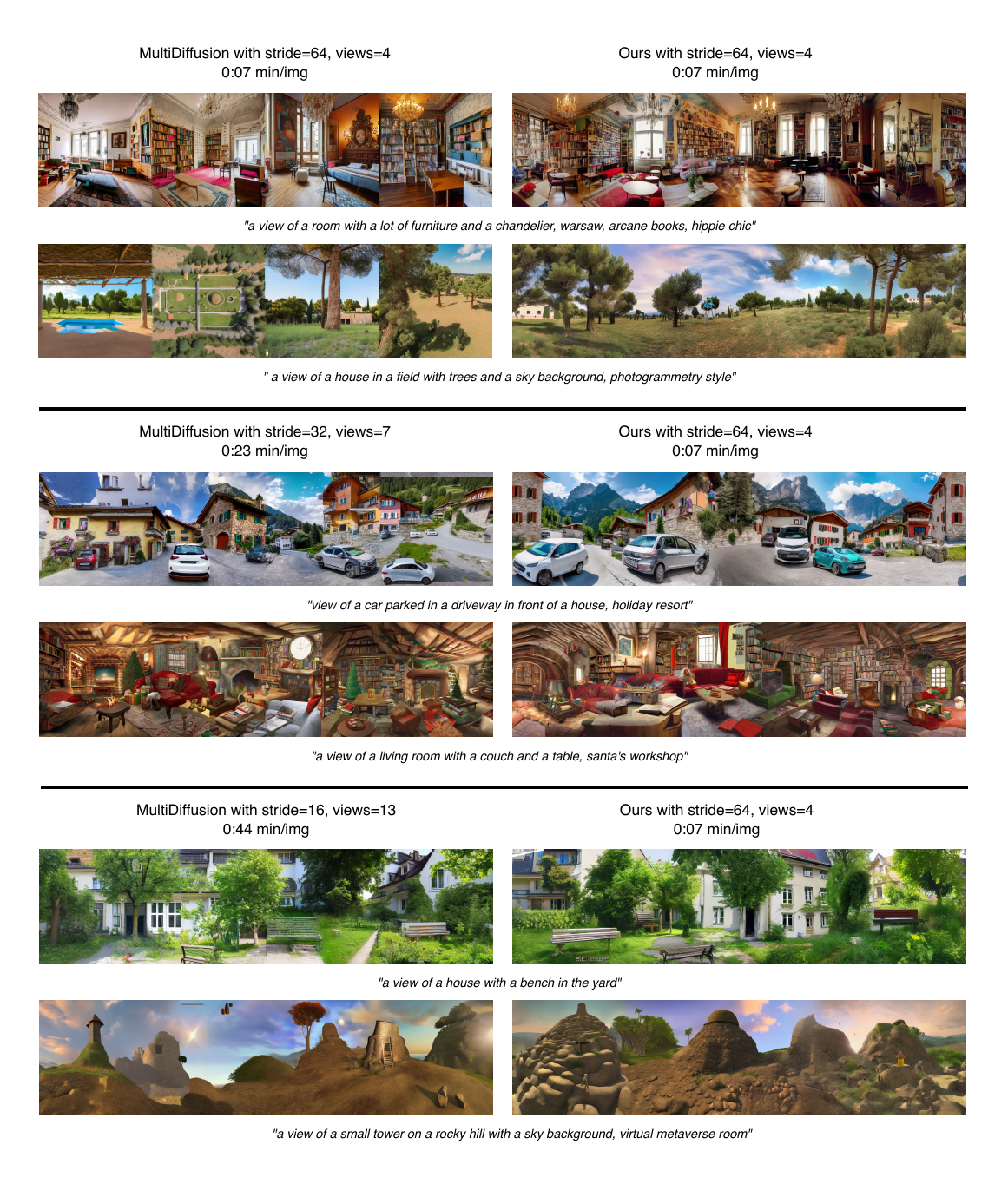}
\caption{
  Comparison of MultiDiffusion \cite{bar2023multidiffusion} with varying stride sizes and our approach.
  MultiDiffusion with stride=64 (no overlap between views) matches our image generation times but produces strong border artifacts and visible seams due to disjoint diffusion paths.
  With stride=32 (50\% overlap), MultiDiffusion still shows visible seams, and only with stride=16 (75\% overlap) does MultiDiffusion produce seamless panoramas, but at the cost of increased computation.
  In contrast, our method consistently achieves seamless panoramas, reducing inference time by 6x without overlapping denoising views, making it more efficient for high-resolution image generation.
  }
\label{fig:md_results} 
\end{figure*}

\begin{figure*}[t]
\centering
\includegraphics[width=0.97\textwidth]{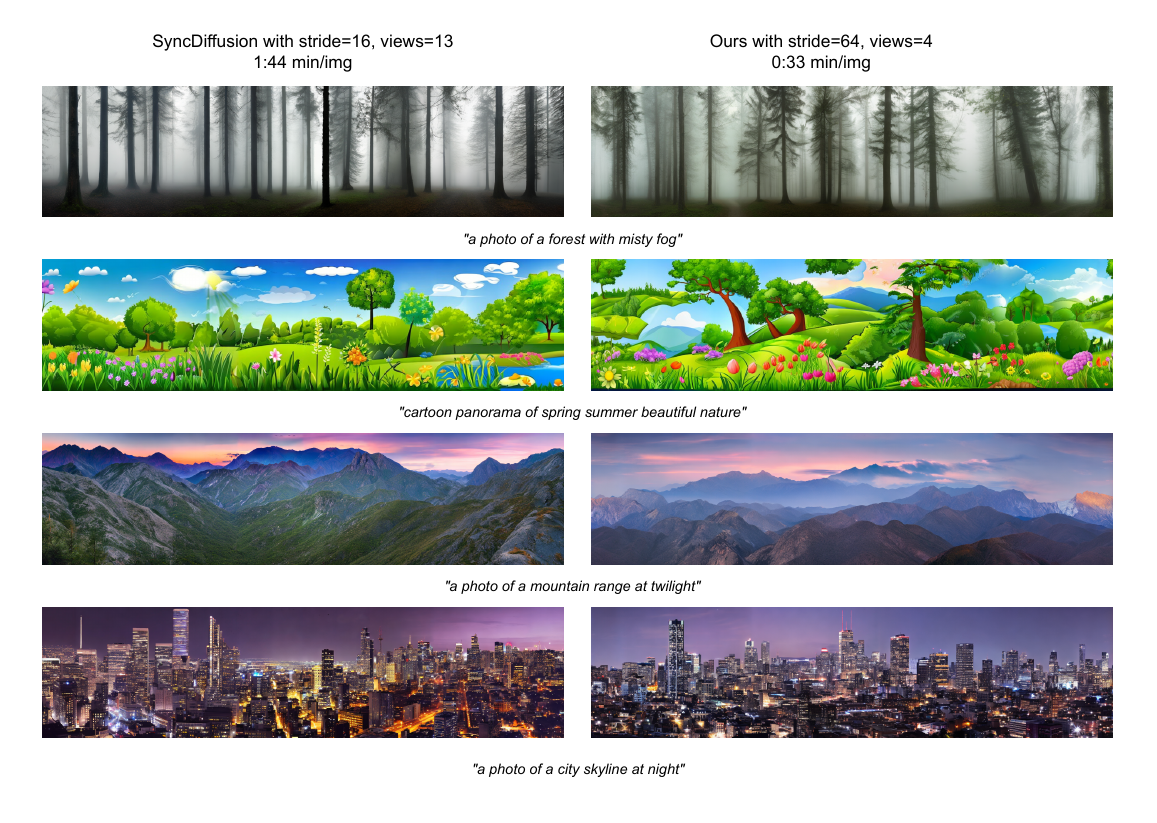}
\caption{
  Our method can also replace the inner MultiDiffusion \cite{bar2023multidiffusion} loop in SyncDiffusion \cite{lee2023syncdiffusion}, leading to a 3x speedup in inference time without noticeable degradation in image quality.
  The generated panorama images are coherent and sharp without border artifacts, demonstrating the effectiveness of our shifted window denoising approach instead of requiring many overlapping patches.
  }
\label{fig:syncdiff_results} 
\end{figure*}

\subsection{Dataset \& Metrics}
To compare with MultiDiffusion \cite{bar2023multidiffusion} and SyncDiffusion \cite{lee2023syncdiffusion}, we follow their evaluation setup which is based on 6 prompts from MultiDiffusion \cite{bar2023multidiffusion}.
We generate 500 panorama images of resolution \(512 \times 2048\) for each prompt and crop each panorama into 6 random \(512 \times 512\) patches for quantitative evaluation.
To evaluate our method and compare it with baselines, we use several metrics.
To measure the image quality and diversity, we use the clean FID \cite{heusel2017gans,parmar2021cleanfid,kynkaanniemi2022role} implementation with CLIP \cite{clip} backbone.
CLIPScore \cite{hessel2021clipscore} measures the alignment between the generated images and the prompts.
Furthermore, we use ImageReward \cite{xu2024imagereward} to evaluate the quality and alignment of the generated images.
ImageReward is a general-purpose text-to-image human preference ranking model, which is trained on a total of 137k pairs of expert comparisons and thus serves as a good proxy to human evaluation.
We report the reward scores for all generated images and their corresponding prompts.
Finally, we measure the time required to generate an image to demonstrate the efficiency of our method and compute the number of windows required to denoise at every timestep.

\section{Results}
\label{sec:results}
This section presents qualitative and quantitative results of our method compared to MultiDiffusion \cite{bar2023multidiffusion}, SyncDiffusion \cite{lee2023syncdiffusion}, and StitchDiffusion \cite{wang2024customizing}.

\subsection{Qualitative Results}  
We start by demonstrating that MultiDiffusion \cite{bar2023multidiffusion} requires large overlap between denoising windows to produce coherent panorama images in \autoref{fig:md_results}.
However, this large overlap results in an increased number of required denoising predictions and thus slow image generation time.
For example, with a stride of 64, which equals the window size, MultiDiffusion \cite{bar2023multidiffusion} produces disjoint diffusion paths and visible seams in the generated images.
Even with a stride of 32, MultiDiffusion \cite{bar2023multidiffusion} still shows visible seams while being 3x slower and requiring 7 instead of 4 denoising windows.
Only with a stride of 16, and thus 13 windows, does MultiDiffusion \cite{bar2023multidiffusion} produce seamless panoramas.
In contrast, our method produces seamless panoramas without overlapping denoising windows, reducing inference time by 6x.

\begin{table*}[t]
  \centering
  \begin{tabular}{lcccccc}
      \toprule
      Model & Stride & Views$\downarrow$ & FID$\downarrow$ & CLIPScore $\uparrow$ & ImageReward$\uparrow$ & Time[min]$\downarrow$ \\
      \midrule
      MultiDiffusion \cite{bar2023multidiffusion}          & 16 & 13 & 3.21 & 31.67 & 0.75 & 0:44  \\
      MultiDiffusion \cite{bar2023multidiffusion}           & 32 & 7 &  3.50 & 31.65 & 0.57 & 0:23  \\
      \rowcolor{Gray} MultiDiffusion \cite{bar2023multidiffusion}           & 64 & 4 & 7.25 & 30.83 & -0.09 & 0:07  \\
      \rowcolor{Gray} Ours                     & 64 & 4 & 3.59 & 31.67 & 0.76 & 0:07  \\
      \midrule
      SyncDiffusion \cite{lee2023syncdiffusion}    & 16 & 13 & 1.86 & 31.85 & 0.62 & 1:44  \\
      SyncDiffusion \cite{lee2023syncdiffusion}   & 32 & 7  & 2.08 & 31.79 & 0.57 & 0:57  \\
      \rowcolor{Gray} SyncDiffusion \cite{lee2023syncdiffusion}   & 64 & 4 & 5.22 & 30.97 & 0.01 & 0:33  \\
      \rowcolor{Gray} Ours            & 64 & 4 & 2.32 & 31.93 & 0.65 & 0:33  \\
      \bottomrule
  \end{tabular}
  \caption{
    Quantitative results on $512 \times 2048$ panorama generation.
    Fast inference with non-overlapping windows comparison is marked in gray.
    Our method consistently produces high-quality images as measured by FID with good image-text alignment as measured by CLIPScore and ImageReward.
    In contrast to the baselines, our method does not require overlapping denoising windows, thus significantly reducing the number of required denoising views and inference time.
    We reach similar performance as the baselines but with a fraction of the time.
    }
  \label{tab:results}
\end{table*}

\begin{figure*}[ht]
  \centering
  \includegraphics[width=0.33\linewidth]{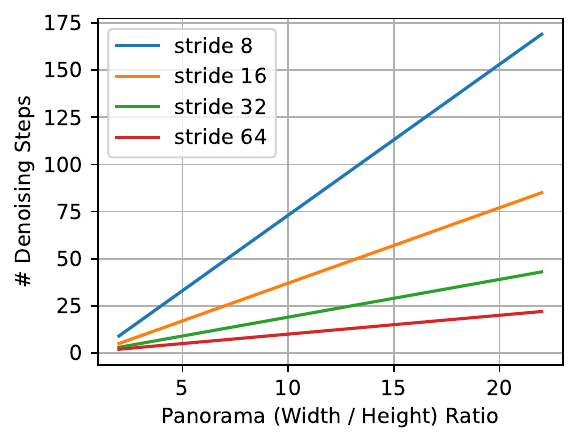}
  \includegraphics[width=0.33\linewidth]{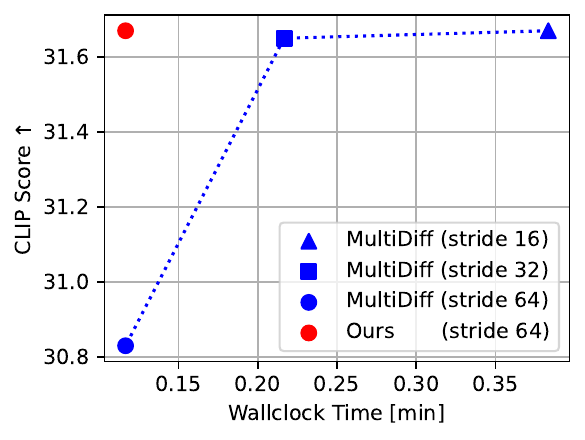}
  \includegraphics[width=0.33\linewidth]{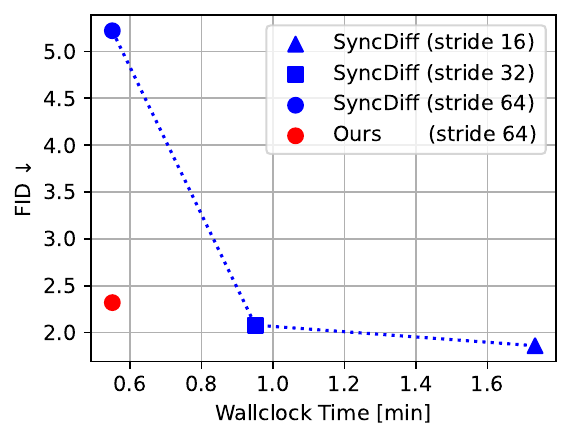}
  \caption{
    \textbf{Left:} The number of total required denoising steps and thus image generation time depends on the stride of denoising windows. Given a default window size of 64, a stride of [64, 32, 16] corresponds to [0\%, 50\%, 75\%] overlap between denoising windows, respectively.
    \textbf{Middle:} CLIPScore comparison of the base StableDiffusion model with MultiDiffusion \cite{bar2023multidiffusion} and our method. As can be seen, our method reaches similar performance as MultiDiffusion \cite{bar2023multidiffusion} but significantly faster.
    \textbf{Right:} FID comparison of SyncDiffusion \cite{lee2023syncdiffusion} with our method. Our method achieves similar FID scores as SyncDiffusion \cite{lee2023syncdiffusion} but with a fraction of the time.
    Notably, our method does not require overlapping denoising windows (window size = stride = 64) and subsequent averaging, making it more efficient.
  }
  \label{fig:quality_vs_time} 
\end{figure*}

In \autoref{fig:syncdiff_results}, we compare our method with SyncDiffusion \cite{lee2023syncdiffusion}.
More specifically, we replace the inner MultiDiffusion \cite{bar2023multidiffusion} loop in SyncDiffusion \cite{lee2023syncdiffusion} with our method to demonstrate the benefit of our approach.
Our method achieves a 3x speedup in inference time without noticeable degradation in image quality.
The panorama images generated by our method are coherent, sharp and without border artifacts, demonstrating the effectiveness of our shifted window denoising approach.

\begin{figure*}[t]
  \centering
  \includegraphics[width=0.97\linewidth]{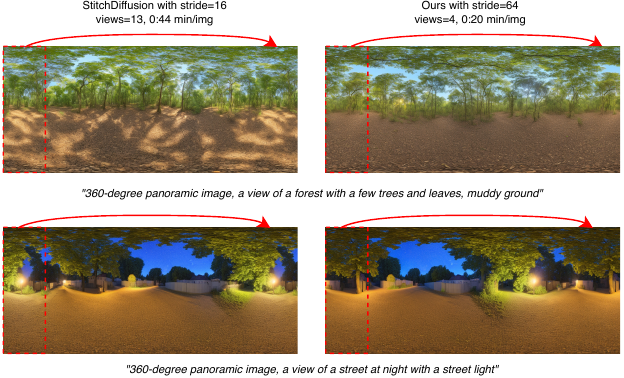}
  \caption{
    Our method can also replace the inner MultiDiffusion \cite{bar2023multidiffusion} loop in StitchDiffusion \cite{wang2024customizing}, a method for spherical image synthesis, leading to a 2x speedup in inference time without noticible degredation in image quality.
    To show the continuity between the leftmost and rightmost sides of the generated image, we copy the left area, indicated by the red dashed box, and paste it onto the rightmost side.
    This demonstrates that both methods produce seamless and plausible 360-degree panoramas based on the input text prompts.
  }
  \label{fig:stitchdiff} 
\end{figure*}

Finally, we use our method to speed up the inference time of StitchDiffusion \cite{wang2024customizing}, see \autoref{fig:stitchdiff}.
Our method produces seamless and plausible 360-degree panoramas based on the input text prompts.
To demonstrate the continuity between the leftmost and rightmost sides of the 360-degree image, we copy the left area and paste it onto the rightmost side.

\subsection{Quantitative Results}
To thoroughly evaluate our method, we compare it with MultiDiffusion \cite{bar2023multidiffusion} and SyncDiffusion \cite{lee2023syncdiffusion} using various metrics.
The results are summarized in \autoref{tab:results} and \autoref{fig:quality_vs_time}.

In \autoref{fig:quality_vs_time}, we show that the number of required denoising steps and thus image generation time depends on the stride of denoising windows.
Next, we compare the CLIPScore of the base StableDiffusion model with MultiDiffusion \cite{bar2023multidiffusion} and our method.
Our method reaches similar performance as MultiDiffusion \cite{bar2023multidiffusion} but significantly faster.
Only with a stride of 16, corresponding to 75\% overlap between denoising windows and roughly 6x more compute time, does MultiDiffusion \cite{bar2023multidiffusion} reach similar CLIPScore.
Finally, we compare the FID scores of SyncDiffusion \cite{lee2023syncdiffusion} with our method.
As already shown in the qualitative results, our method achieves similar image quality in terms of FID scores as MultiDiffusion \cite{bar2023multidiffusion} and SyncDiffusion \cite{lee2023syncdiffusion} but with a fraction of the time.
Interestingly, our method achieves better image-text alignment as measured by CLIPScore and ImageReward while being significantly faster.
We hypothesize this is due to reduced following of the text guidance by averaging separate overlapping denoising predictions within each timestep.

\section{Limitations \& Future Work}
\label{sec:limitations}
Our method relies on random shifts of non-overlapping denoising windows over time to ensure coherent transitions across the panorama image.
While this approach significantly reduces the computational complexity of the denoising process and speeds up image generation time, it does not guarantee the same level of image quality as measured by FID as dense MultiDiffusion \cite{bar2023multidiffusion} with a very small stride of 16.
In future work, we plan to investigate the impact of dynamically adjusted different stride sizes during the denoising process on image quality and generation time to combine the benefits of both methods.

\section{Conclusion}
\label{sec:conclusion}
This paper presents a novel approach for coherent high-resolution image generation using diffusion models.
Instead of relying on overlapping denoising windows, our method shifts the denoising windows over time, ensuring that seams in one timestep are corrected in the next.
This results in coherent, high-resolution images with fewer overall steps, enhancing the quality and efficiency of the generation process.
We demonstrate the effectiveness of our approach through qualitative and quantitative evaluations, comparing it with MultiDiffusion \cite{bar2023multidiffusion}, SyncDiffusion \cite{lee2023syncdiffusion}, and StitchDiffusion \cite{wang2024customizing}.
Our method consistently produces high-quality images with good image-text alignment while significantly reducing the number of required denoising views and inference time.
By eliminating the need for overlapping denoising windows, our approach offers a more efficient and practical solution for high-resolution image generation, advancing the state of the art in the field.

\section{Societal Impact}
Generative image models can be misused to create deepfakes, infringe on copyrights, and produce biased images.
These problems can lead to fake news, privacy invasion, and reinforcing stereotypes.
Our method builds on existing foundational models and thus shares these risks.
To reduce them, it's important to develop better deepfake detection, protect intellectual property, and follow ethical guidelines.
Labeling synthetic content and learning to use generative models responsibly is essential.
By focusing on these protective measures, we can use generative models effectively while minimizing their negative impacts.

\section*{Acknowledgements}
This work was supported by the BMBF project SustainML (Grant 101070408).

{\small
\bibliographystyle{ieee_fullname}
\bibliography{egbib}
}

\end{document}